\documentclass[conference, 10pt]{IEEEtran}

\renewcommand\baselinestretch{0.99}

\usepackage[labelsep=period]{caption}

\usepackage{cite}

\usepackage{soul} 
\usepackage{xcolor}

\usepackage{times}

\usepackage[hidelinks, bookmarks=false]{hyperref}

\usepackage{booktabs}

\usepackage{cmap}
\usepackage{mmap}

\usepackage{lipsum}

\usepackage[]{graphicx}

\usepackage[font=small]{subfig}

\usepackage{amsmath}
\usepackage{amsfonts}

\usepackage[T1]{fontenc}

\usepackage{balance}

\usepackage{siunitx}

\usepackage{enumitem}

\usepackage{algorithm}
\usepackage{algpseudocode}

\begin{document}
	
	\title{A Hardware-Friendly Algorithm for Scalable Training and Deployment of Dimensionality Reduction Models on FPGA}

	\author{\IEEEauthorblockN{Mahdi Nazemi, Amir Erfan Eshratifar, and Massoud Pedram}
	\IEEEauthorblockA{Department of Electrical Engineering, University of Southern California, Los Angeles, CA, USA\\
		\url{{mnazemi, eshratif, pedram}@usc.edu}}
	}
		
	\maketitle
	\begin{abstract}
		With ever-increasing application of machine learning models in various domains such as image classification, speech recognition and synthesis, and health care, designing efficient hardware for these models has gained a lot of popularity.
		While the majority of researches in this area focus on efficient deployment of machine learning models (a.k.a inference), this work concentrates on challenges of training these models in hardware.
		In particular, this paper presents a high-performance, scalable, reconfigurable solution for both training and deployment of different dimensionality reduction models in hardware by introducing a hardware-friendly algorithm.
		Compared to state-of-the-art implementations, our proposed algorithm and its hardware realization decrease resource consumption by 50\% without any degradation in accuracy.
	\end{abstract}

	\section{Introduction}
	Advancements in developing high-performance hardware platforms like GPUs have been a significant enabler for shifting machine learning models, such as neural networks, from rather theoretical concepts to practical solutions to a wide variety of problems.
	For example, different applications available on smartphones use machine learning models in order to perform services such as categorizing photos based on faces present in the picture through face recognition, digital assistance through speech recognition and synthesis, suggesting similar applications through recommendation systems, and so forth.
	While existing hardware platforms like GPUs are capable of performing these tasks, designing more energy-efficient and high-performance hardware is crucial in order to allow pervasive deployment of machine learning models across different platforms, from  data centers to smartphones and the Internet of Things (IoT) devices.

	Even tough there has been a substantial effort to design accelerators or use alternative methods for efficient deployment of machine learning models such as convolutional neural networks \cite{lin2018fft}, the training phase of these models has been overlooked.
	One of the difficulties in designing hardware that is capable of training is that the training phase is typically much more complicated and computationally expensive compared to inference.
	We believe that designing high-performance and/or energy-efficient hardware for training is of high importance due to several reasons.
	First, transferring users' data to remote servers puts the users' privacy in danger.
	Second, communication latency may affect latency-critical applications like online control systems.
	Third, adapting to a changing environment requires updating model parameters frequently and may be costly, especially in bandwidth-limited devices.
	And lastly, the energy consumption of wireless modules that need to send/receive data to/from cloud is relatively high \cite{altamimi2015energy}.
	A suitable hardware for training machine learning models should operate with high performance, be scalable, and be able to train various models by exploiting resource sharing and real-time reconfiguration.

	Dimensionality reduction, which is the target machine learning model in this paper, has several advantages.
	First, it removes redundant information from the set of input features, which typically improves the performance of machine learning models.
	Features that are highly correlated or are closely dependent do not carry much additional information and only make the model more complicated and computationally expensive.
	Second, transforming features into a lower-dimensional space is more suitable from a hardware design point of view since it leads to a less complex design, less resource consumption, lower memory usage, and so on.

	This work presents a hardware-friendly algorithm for high-performance, scalable, and reconfigurable training and deployment of dimensionality reduction models.
	The main focus of this work is to deal with scalability issue of existing hardware implementations for dimensionality reduction while it also considers reconfigurability in order to use the same hardware for various dimensionality reduction algorithms.
	The rest of this paper is organized as follows.
	Section~\ref{sec:related} reviews prior work in designing hardware for dimensionality reduction.
	Section~\ref{sec:prelim} provides some background information about the effect of dimensionality reduction on other machine learning models, explains specific algorithms for dimensionality reduction, and discusses the scalability issue of existing implementations.
	Section~\ref{sec:implementation} presents our proposed algorithm for dimensionality reduction as well as its hardware implementation.
	Section~\ref{sec:experiments} demonstrates experimental results and finally, Section~\ref{sec:conclusion} concludes the paper.

	\section{Related Work} \label{sec:related}
	One of the most successful attempts at designing an algorithm for scalable dimensionality reduction is random projection.
	Random projection is based on the Johnson-Lindenstrauss lemma \cite{johnson1984extensions} and is much simpler than other distance-preserving algorithms such as PCA (Principal Component Analysis).
	Random projection has been applied to various applications and it has been shown that its quality of results are comparable to other algorithms \cite{bingham2001random, fern2003random, dasgupta2000experiments}.
	Recently, Fox~\textit{et al.} \cite{fox2016random} implemented random projection in hardware using a simple algorithm that only requires addition and subtraction.
	The shortcoming of the random projection though, is that it only deals with mixture of Gaussian variables.
	In other words, it only considers second-order statistics when transforming data points to a lower-dimensional space.

	In order to consider higher-order statistics (HOS), another class of algorithms known as ICA (Independent Component Analysis) \cite{comon1994independent} is used.
	Among different algorithms that implement ICA, EASI (Equivariant Adaptive Separation via Independence) \cite{cardoso1996equivariant} has been one of the most suitable ones from a hardware implementation standpoint because it only requires addition and multiplication.
	EASI includes both training and inference of a dimensionality reduction model that implements ICA.
	Meyer-Baese \textit{et al.} \cite{meyer2015independent} implement EASI in hardware, but their work has a few shortcomings.
	First, the clock frequency and throughput are very low.
	Second, the clock frequency decreases by increasing the number of input or output dimensions.
	This is a serious problem, especially given the high dimensionality of datasets in existing machine learning problems.
	Nazemi \textit{et al.} \cite{nazemi2017high} try to address these issues by defining a new approximation to stochastic gradient descent algorithm that is suitable for hardware implementation.
	Their implementation of EASI using the aforementioned approximation increases the clock frequency by one order of magnitude compared to \cite{meyer2015independent} and keeps the clock frequency independent of input and output dimensions.
	However, \cite{nazemi2017high} suffers from poor scalability in that its hardware implementation for four input dimensions and two output dimensions consume more than one third of the digital signal processing (DSP) blocks on their target FPGA platform.
	In general, the major problem with PCA and ICA is their high hardware complexity in terms of adders and multipliers, which limits their scalability to larger dimensions.

	\section{Preliminaries} \label{sec:prelim}
	This section demonstrates the effect of dimensionality reduction on the accuracy of other machine learning models and provides background information about some of the dimensionality reduction models and scalability issue of existing hardware implementations.
	Additionally, it provides a short mathematical description of the EASI algorithm that is used later in Section~\ref{sec:implementation} for justifying the proposed algorithm.

	\subsection{Effect of Dimensionality Reduction on Accuracy}
\begin{figure}[t]
		\centering
		\subfloat[MNIST Dataset - Image] {
			\includegraphics[width=0.8\columnwidth]{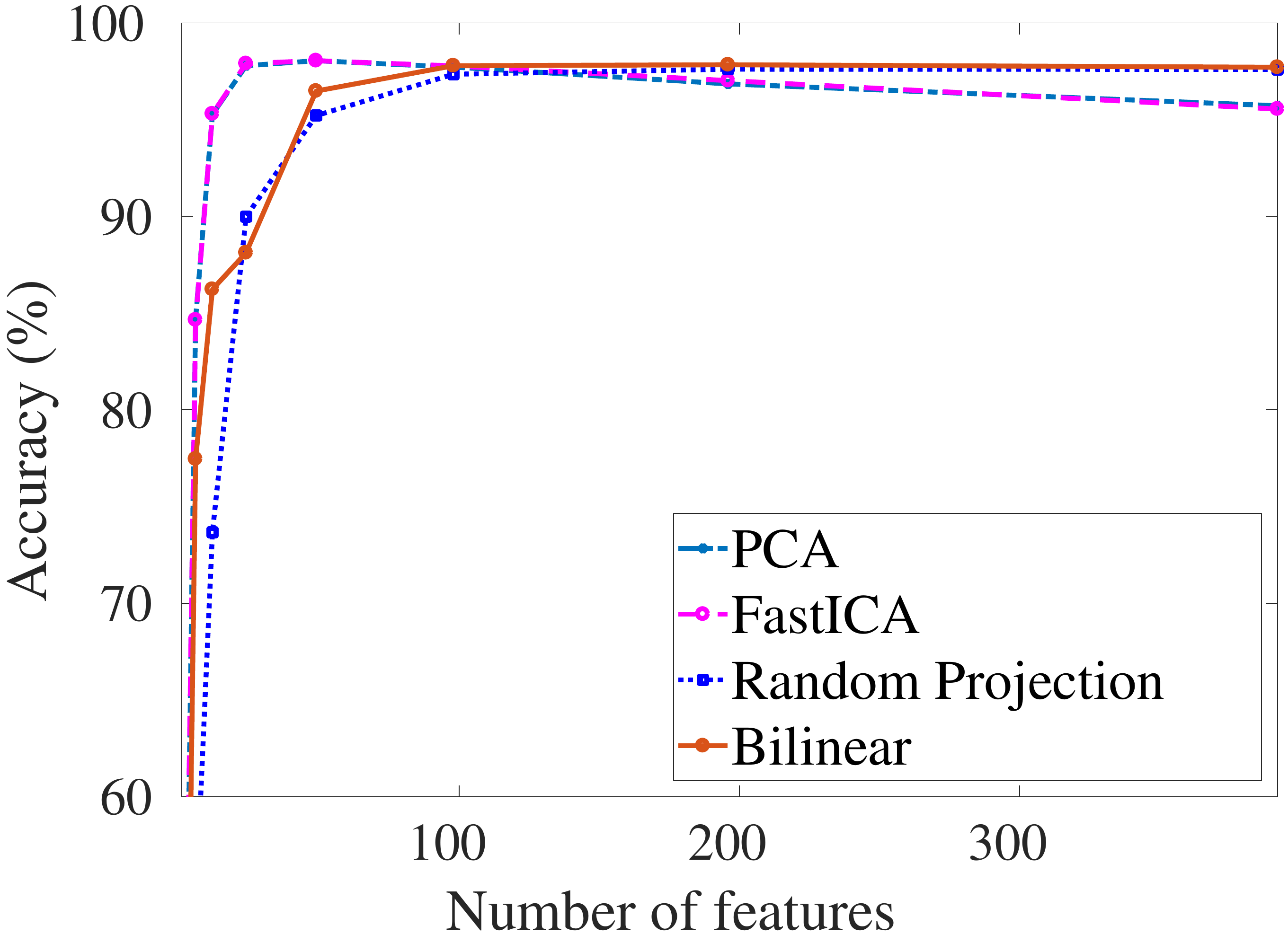}
			\label{fig:mnist}
			}
			
		\subfloat[HAR Dataset - Time Series] {
			\includegraphics[width=0.8\columnwidth]{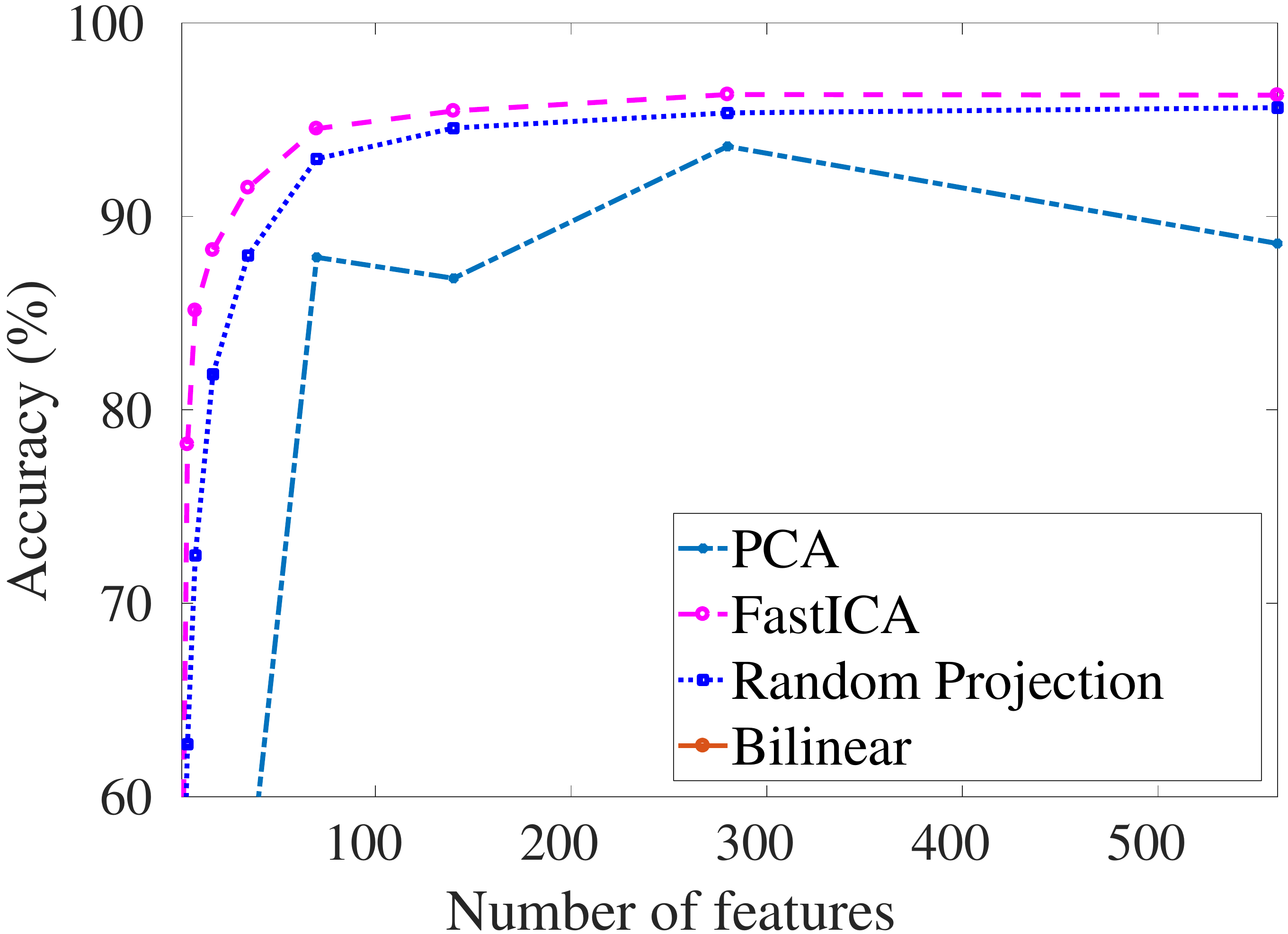}
			\label{fig:har}
			}
			
		\subfloat[Ads Dataset - Natural Language and Image] {
			\includegraphics[width=0.8\columnwidth]{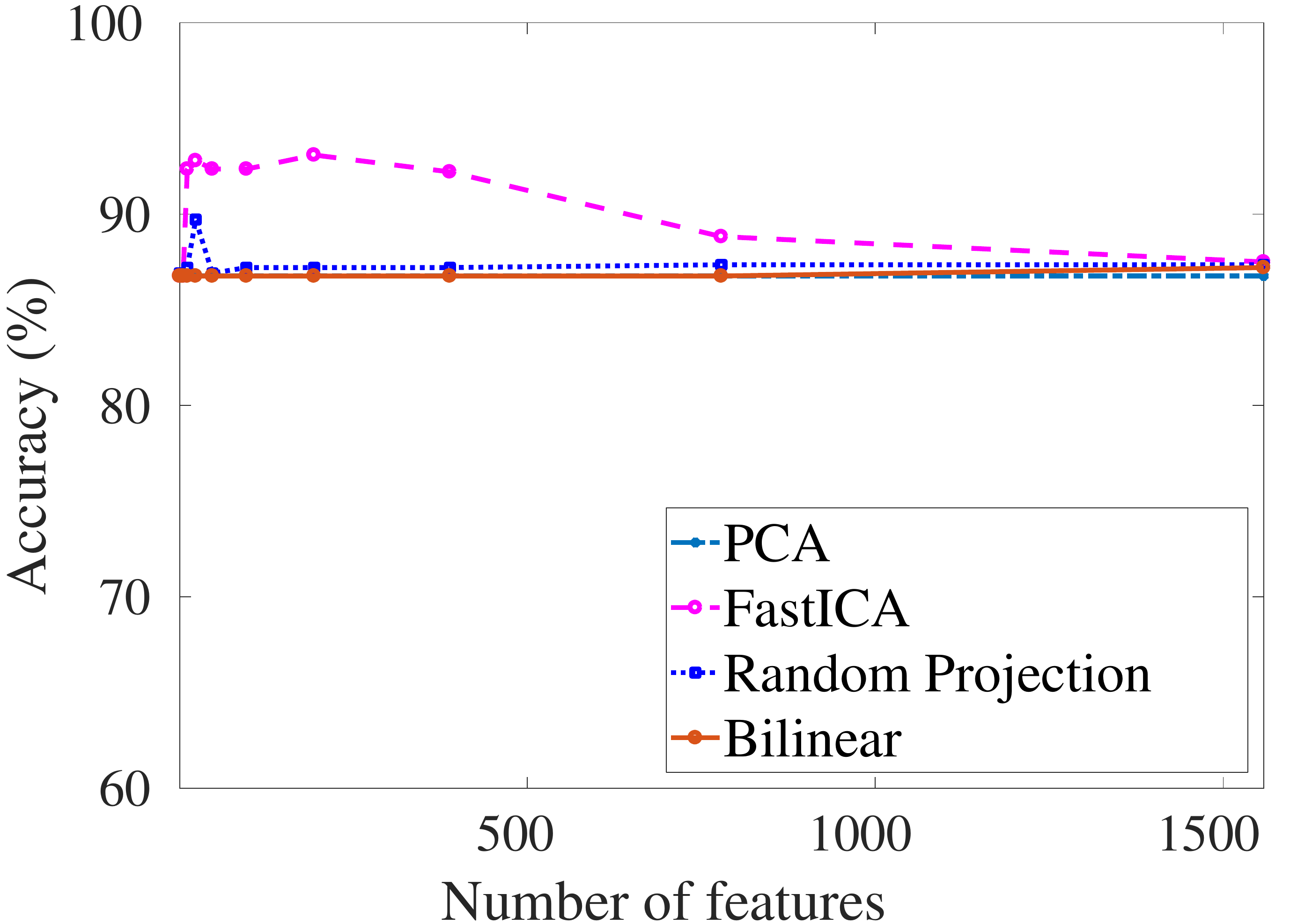}
			\label{fig:ads}
			}
			
		\caption{Classification accuracy for various datasets, different dimensionality reduction algorithms, and different number of input features.}
		\label{fig:dim-accuracy}	
	\end{figure}

	One of the major advantages of dimensionality reduction is that it decreases both computation and storage complexity.
	When a dimensionality reduction algorithm is applied, the number of input features is reduced and therefore, a smaller amount of memory is required to store the input features.
	Additionally, the machine learning model that follows the dimensionality reduction model needs to deal with a lower number of input features, which in turn makes that model less computationally expensive.
	Fig.~\ref{fig:dim-accuracy} compares the classification accuracy for various datasets, different dimensionality reduction algorithms, and different number of input features.
	For all these datasets, an artificial neural network with two hidden layers is trained in order to perform classification.

	Fig.~\ref{fig:mnist} demonstrates the effect of reducing dimensionality of input features on the classification accuracy for images in the MNIST dataset \cite{lecun2010mnist}.
	The MNIST database of handwritten digits includes 70,000 samples where each sample is a 28x28 image (784 pixels total) and the objective is to classify each sample into one of ten classes 0-9.
	It can be observed that reducing the number of input features to about 100 (\char`\~ 8x reduction) using random projection and bilinear transform does not affect the classification accuracy.
	In this dataset, PCA and ICA can achieve even higher degrees of reduction (\char`\~ 16x) without any noticeable accuracy degradation.

	Similarly, Fig.~\ref{fig:har} shows the effect of dimensionality reduction on HAR dataset \cite{anguita2013public}.
	This dataset uses the accelerometer and gyroscope embedded in a smartphone to measure a group of volunteers' activities over a period of time and the objective is to classify each sample into one of six classes that determines a volunteer's activity.
	The original number of input features for each sample is 561.
	It can be seen that ICA and random projection outperform the other two methods and can achieve about 6x reduction in input features without significantly affecting the classification accuracy.
	Bilinear transform does not perform well in this dataset and the classification accuracy is below 60\%.

	Lastly, Fig.~\ref{fig:ads} demonstrates the effect of dimensionality reduction on Ads dataset \cite{kushmerick1999learning}.
	This dataset represents a set of possible advertisements on Internet pages where each sample has 1558 input features, which include the geometry of the image, phrases occurring in the URL, the anchor text, words occurring near the anchor text, etc., and the objective is to determine whether a sample is an advertisement or not.
	It is observed that reducing the number of input features to five (\char`\~ 300x reduction) does not affect the classification accuracy.
	This observation corroborates the results presented in prior work on this dataset.

	It can be concluded that dimensionality reduction models typically perform well for a variety of datasets, including the ones that deal with images, time series, and/or natural language.
	However, various dimensionality reduction algorithms perform differently on these datasets.
	This is another reason why a piece of hardware that is capable of implementing different dimensionality reduction algorithms is superior.

	\subsection{Random Projection}
	In a random projection, the lower-dimensional input features are found by multiplying the original input features by a randomly generated matrix $R$:
	\begin{equation}
		{\mathbf{v}}_{n \times 1} = R_{n \times m}{\mathbf{x}}_{m \times 1} \quad m \geq n
		\label{eq:random-projection}
	\end{equation}	
	where $\mathbf{x}$ is a column vector of input features, $R$ is the randomly generated matrix, $\mathbf{v}$ is a column vector of  features in the lower-dimensional space, $m$ is the dimensionality of input features, and $n$ is the dimensionality of features in the lower-dimensional space.
	The elements of $R$ (i.e. $r_{ij}$) are often sampled from a Gaussian distribution, however, there are other proposed distributions such as the ones introduced in \cite{achlioptas2001database, li2006very} that are more suitable for hardware implementation.
	In this work, we use the distribution that is described in \cite{fox2016random}:
	\begin{equation*}
		r_{ij} = 
		\begin{cases}
			1,     & \text{with probability} \quad 1/(2n) \\
			0,     & \text{with probability} \quad 1 - 1/n \\
			-1,    & \text{with probability} \quad 1/(2n)
		\end{cases}
		\label{eq:distribution}
	\end{equation*}	
	The advantage of this distribution is that is replaces all multiplications with addition and subtraction and therefore, reduces the hardware cost.

	One of the major advantages of random projection is that the model does not need to be trained based on input data.
	As a result, the $R$ matrix can be computed offline without having any information about upcoming input features.

	\subsection{PCA Whitening}
	PCA whitening is an important preprocessing step in many machine learning algorithms.
	The objective of PCA whitening is to transform data to a lower-dimensional space such that features are less correlated with each other and all features have the same variance (typically unit variance).
	This can be written as
	\begin{equation}
	{\mathbf{z}}_{n \times 1} = W_{n \times m}{\mathbf{x}}_{m \times 1} \quad m \geq n
	\label{eq:pca-whitening}
	\end{equation}
	where $\mathbf{x}$ is a column vector of input features, $W$ is the whitening matrix, $\mathbf{z}$ is a column vector of whitened features, $m$ is the dimensionality of input features, and $n$ is the dimensionality of features in the lower-dimensional space.
	
	\subsection{EASI Algorithm}
	ICA can be defined as a generative model in which input features are modeled as linear combinations of some independent components:
	\begin{equation*}
	{\mathbf{x}}_{m \times 1} = A_{m \times n}{\mathbf{s}}_{n \times 1} \quad m \geq n
	\end{equation*}
	where $\mathbf{x}$ is a column vector of input features, $A$ is the mixing matrix, $\mathbf{s}$ is a column vector of random independent components, $m$ is the dimensionality of input features, and $n$ is the dimensionality of independent components in the lower-dimensional space.
	The objective of ICA is to estimate the mixing matrix and independent components without having any prior information about them.

\begin{figure*}[t]
		\centering
		\subfloat[] {
			\includegraphics[width=0.3\textwidth]{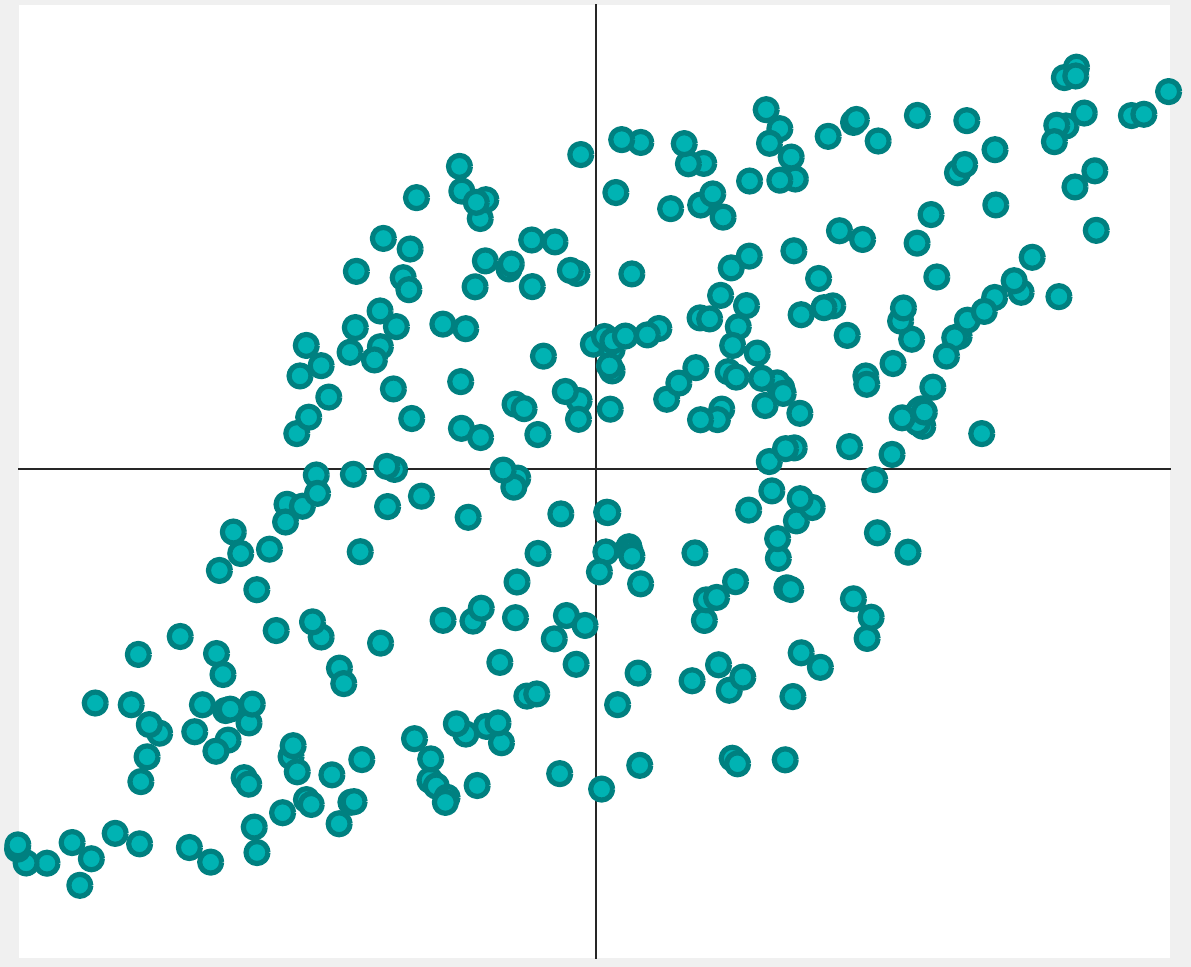}
			\label{fig:transformed}
		} \quad
		\subfloat[] {
			\includegraphics[width=0.3\textwidth]{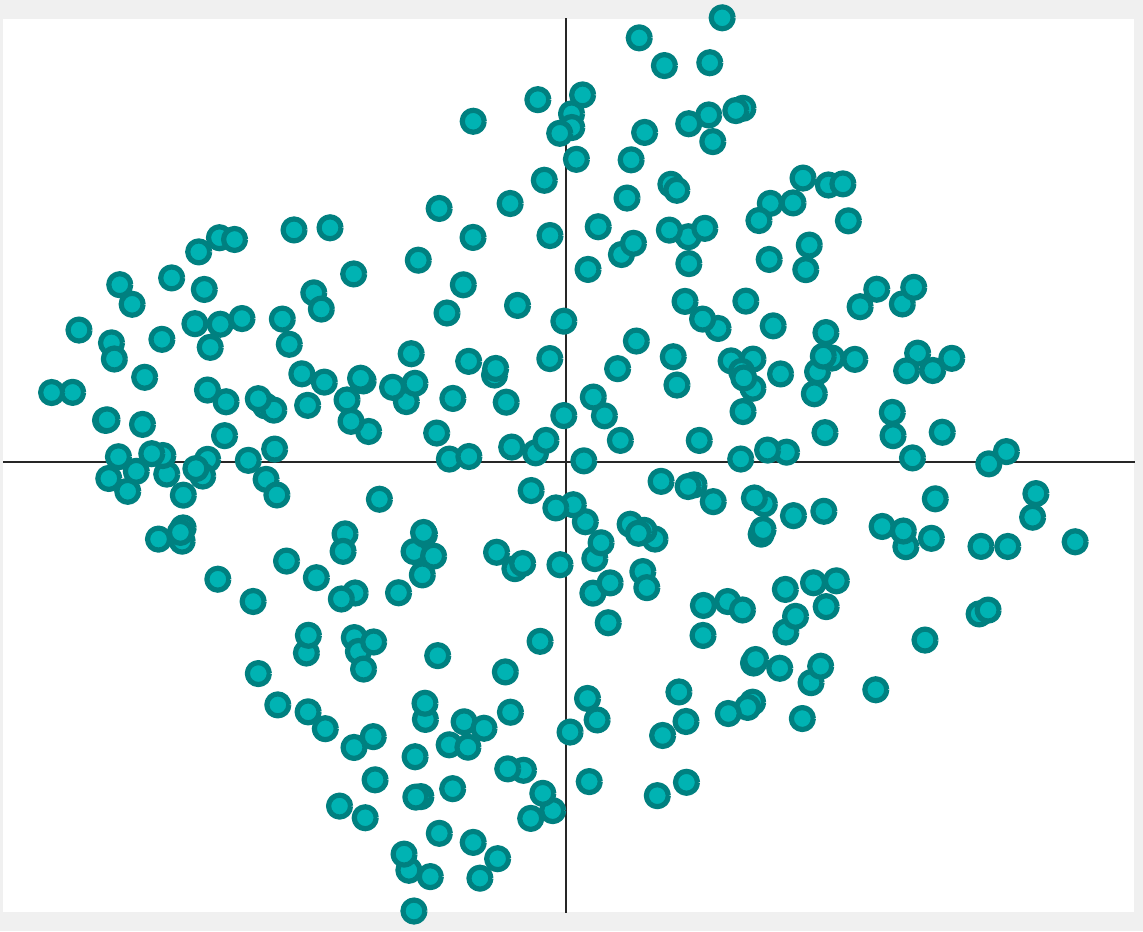}
			\label{fig:whitened}
		} \quad
		\subfloat[] {
			\includegraphics[width=0.3\textwidth]{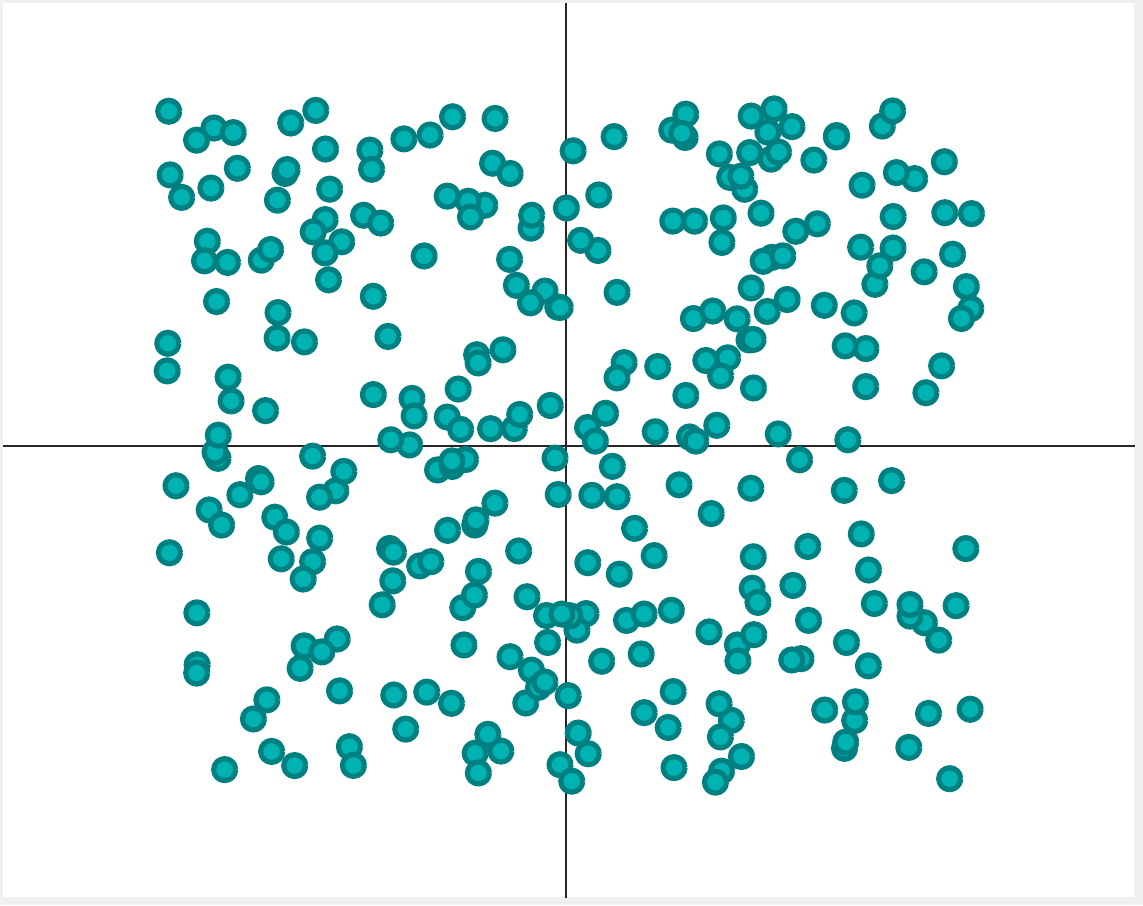}
			\label{fig:rotated}
		}
		
		\caption{Finding independent components by whitening (b) followed by a rotation (c). The whitened features have the same variance and therefore, an appropriate rotation can find the original independent components.}
		\label{fig:whitening-rotation}	
	\end{figure*}

	The estimation can be achieved by applying a whitening matrix followed by an orthogonal transformation, i.e. rotation, of intermediate input features.
	This process is illustrated in Fig.~\ref{fig:whitening-rotation}.
	The whitening step can be written as
	\begin{equation*}
		{\mathbf{z}}_{n \times 1} = W_{n \times m}{\mathbf{x}}_{m \times 1} \quad m \geq n
	\end{equation*}
	where $\mathbf{x}$ is a column vector of input features, $W$ is the whitening matrix, and $\mathbf{z}$ is a column vector of whitened features.
	By definition, $W$ is a whitening matrix if $\mathbf{z}$ is spatially white, that is
	\begin{equation*}
		\Sigma_{{\mathbf{z}}_{n \times n}} = {\mathrm{E}}[{\mathbf{z}}{\mathbf{z}}^T] = I_{n \times n}
	\end{equation*}	
	where $\Sigma_{{\mathbf{z}}}$ is the covariance matrix of ${\mathbf{z}}$, and ${\mathrm{E}}$ is the expectation operator.

	One of the methods for finding the whitening matrix is to minimize the the Kullback-Leibler divergence \cite{kullback1951information} between $\Sigma_{{\mathbf{z}}}$ and $I$.
	The adaptive updating algorithm for $W$ that minimizes Kullback-Leibler divergence can be written as
	\begin{equation}
		W_{k + 1} = W_{k} - \mu_{k}[{\mathbf{z}_{k}}{\mathbf{z}_{k}}^T - I] W_{k}
		\label{eq:whitening}
	\end{equation}	
	in which $k$ is the iteration index and $\mu$ is the learning rate.
	The learning rate does not necessarily need to change across iterations, i.e. $\mu_{k} = \mu$.	

	The goal of EASI is to find a separation matrix that provides an estimate of independent components without having any prior information about independent components, $\mathbf{s}$, or the mixing matrix, $A$.
	This can be written as
	\begin{equation}
	{\mathbf{y}}_{n \times 1} = B_{n \times m}{\mathbf{x}}_{m \times 1}
	\label{eq:output-features}
	\end{equation}
	where $\mathbf{y}$ is a column vector of estimates of independent components and $B$ is the separation matrix.

	The separation matrix can be found by applying the whitening matrix followed by a rotation, i.e.
	\begin{equation*}
	B_{n \times m} = U_{n \times n} W_{n \times m}
	\end{equation*}
	where $U$ is an orthogonal matrix.
	An adaptive updating algorithm that keeps $U$ an orthogonal matrix can be found by
	\begin{equation}
	U_{k + 1} = U_{k} - \mu_{k}[{\mathbf{g}}({\mathbf{y}_{k}}){\mathbf{y}_{k}}^T - {\mathbf{y}_{k}}{\mathbf{g}}({\mathbf{y}_{k}})^T] U_{k}
	\label{eq:rotation}
	\end{equation}
	where ${\mathbf{g}}(.)$ is a nonlinear function that introduces HOS into the problem.

	A global adaptive updating algorithm for the separation matrix can be found by $B_{k + 1} = U_{k + 1} W_{k + 1}$ and plugging in $W_{k + 1}$ and $U_{k + 1}$ from Eq.~\ref{eq:whitening} and Eq.~\ref{eq:rotation}, respectively.
	By neglecting the $\mu^{2}$ term, the adaptive updating algorithm for $B$ can be written as
	\begin{equation}
	B_{k + 1} = B_{k} - \mu_{k}[{{\mathbf{y}_{k}}{\mathbf{y}_{k}}^T - I + \mathbf{g}}({\mathbf{y}_{k}}){\mathbf{y}_{k}}^T - {\mathbf{y}_{k}}{\mathbf{g}}({\mathbf{y}_{k}})^T] B_{k}
	\label{eq:separation}
	\end{equation}
	Eq.~\ref{eq:separation} is known as the EASI algorithm for independent component analysis.

	\subsection{Scalability Problem}
\begin{figure*}[bt]
		\centering
		\includegraphics[width=0.8\textwidth]{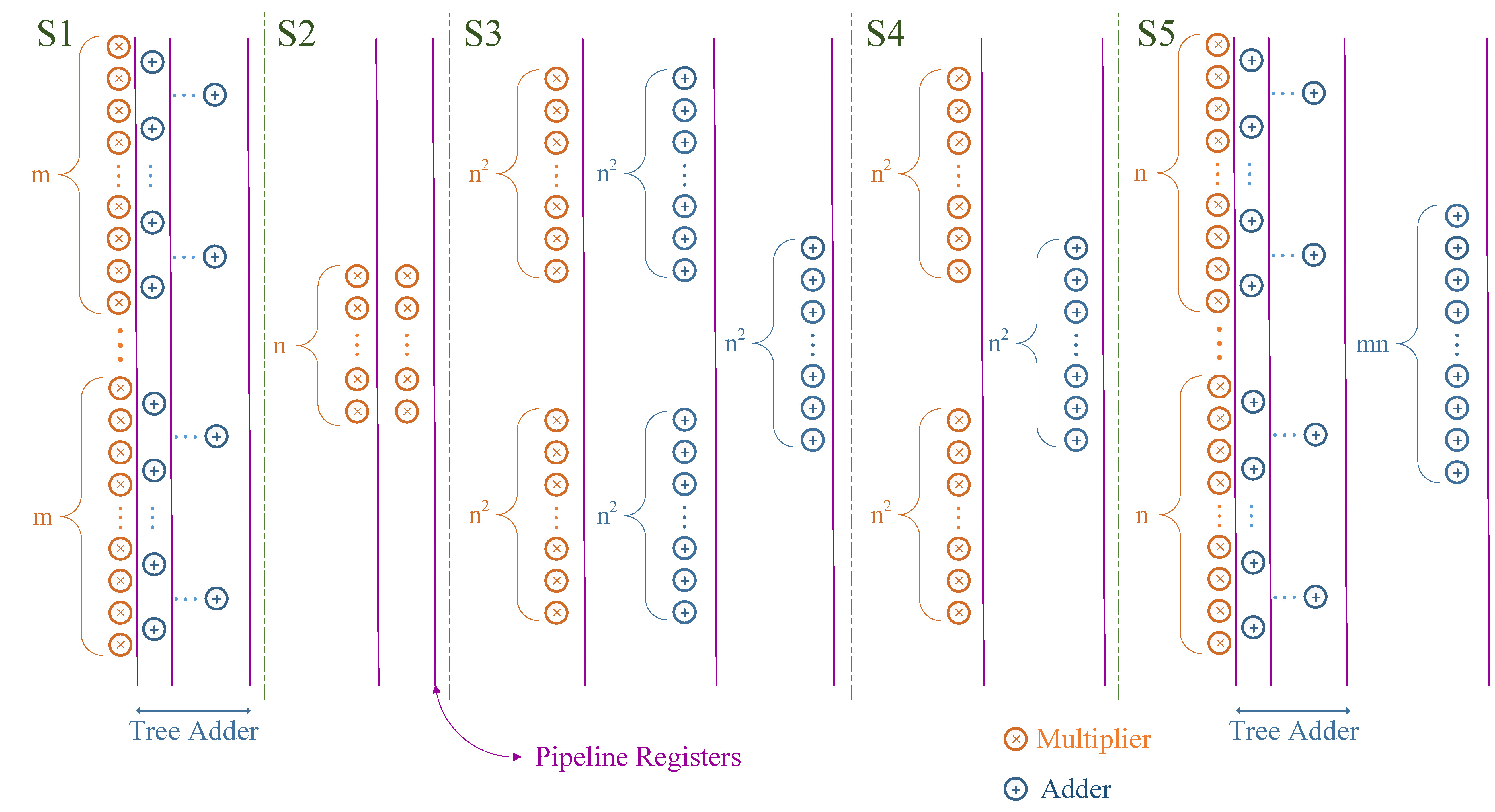}
		\caption{Hardware implementation of the EASI algorithm.}
		\label{fig:easi-smbgd}	
	\end{figure*}

	Although the hardware implementation of EASI algorithm that is presented in \cite{nazemi2017high} increases the clock frequency by an order of magnitude compared to its prior work, the design suffers from poor scalability.
	Figure~\ref{fig:easi-smbgd} depicts different stages of hardware implementation of EASI algorithm based on \cite{nazemi2017high}.
	The algorithm consists of five high-level stages where each stage is responsible for one of the steps explained in Algorithm~\ref{alg:easi-smbgd}.
	By calculating the number of adders and multipliers required for implementing each stage, one can observe that the hardware complexity of both adder and multiplier units is $\mathcal{O}(mn^2)$.
	This is obviously not a scalable algorithm and its hardware implementation will occupy the resources available on an FPGA very quickly.

	\begin{algorithm}[tb]
		\caption{EASI with Modified Update Rule}
		\label{alg:easi-smbgd}
		
		\begin{algorithmic}[1]
			\Require
			\Statex $\mathbf{x}$: \text{input features}	
				
			\Ensure
			\Statex $B$: \text{separation matrix}
			\Statex $\mathbf{y}$: \text{input features in the lower-dimensional space}
			
			\Repeat
				\State Update $\mathbf{y}$ according to Eq.~\ref{eq:output-features}
				\State Apply cubic nonlinearity to $\mathbf{y}$
				\State Calculate ${{\mathbf{y}_{k}}{\mathbf{y}_{k}}^T - I + \mathbf{g}}({\mathbf{y}_{k}}){\mathbf{y}_{k}}^T - {\mathbf{y}_{k}}{\mathbf{g}}({\mathbf{y}_{k}})^T$
				\State Update relative gradient
				\State Update separation matrix according to Eq.~\ref{eq:separation}
			\Until{convergence}
		\end{algorithmic}
	\end{algorithm}	

	\section{Proposed Solution} \label{sec:implementation}
	By looking at Eq.~\ref{eq:separation} carefully, we observe that the ${\mathbf{y}_{k}}{\mathbf{y}_{k}}^T - I$ term is in fact responsible for dealing with second-order statistics while the $\mathbf{g}({\mathbf{y}_{k}}){\mathbf{y}_{k}}^T - {\mathbf{y}_{k}}{\mathbf{g}}({\mathbf{y}_{k}})^T$ term deals with higher-order statistics.
	Adding this to what was explained in Fig.~\ref{fig:whitening-rotation} and the fact that random projection is a suitable algorithm for dealing with second-order statistics, we propose the following solution for implementing the EASI algorithm more efficiently.
	Initially, input features are fed to a random projection module that preserves the second-order distance among these features, but reduces the dimensionality to an intermediate value $p$.
	After that, a modified datapath for the EASI algorithm that bypasses the ${\mathbf{y}_{k}}{\mathbf{y}_{k}}^T - I$ term applies a rotation to intermediate input features in order to find features that are independent and have a dimensionality $n$.
	This process is repeated until the model is trained and can be used later for inference.
	The major advantage of this solution is that the number of inputs to the EASI module will be decreased because of the dimensionality reduction that is performed by the random projection module.
	This in turn reduces the number of adders and multipliers in the EASI module linearly due to the linear dependency between the hardware complexity and number of input dimensions.

	Given the fact that the hardware implementation of random projection has a low overhead, this enables dealing with higher number of input dimensions at the cost of slightly increasing latency.
	The increase in latency is due to the fact that EASI applies whitening and rotation in parallel, but the proposed solution applies whitening and rotation sequentially.
	However, the asymptotic latency of random projection is negligible compared to EASI and can be ignored.

	By comparing Eq.~\ref{eq:pca-whitening} with Eq.~\ref{eq:output-features} and Eq.~\ref{eq:whitening} with Eq.~\ref{eq:separation}, we observe that the algorithm's flow for implementing PCA whitening and EASI are the same.
	In both algorithms, not only the basic operations like matrix-vector multiplication, matrix-matrix multiplication, and update rule are the same, but also the dimensionality of matrices and vectors are the same.
	As a result, a hardware that implements EASI for ICA can be used for implementing PCA whitening as well.
	The only difference is that EASI has an additional term $\mathbf{g}({\mathbf{y}_{k}}){\mathbf{y}_{k}}^T - {\mathbf{y}_{k}}{\mathbf{g}}({\mathbf{y}_{k}})^T$ that needs to be bypassed for PCA whitening simply by using a multiplexer.
	This allows real-time reconfigurability by issuing proper control signals for each algorithm and therefore, enables using the same hardware for both PCA whitening and ICA.

	In conclusion, our hardware implementation will comprise of a random projection module followed by an EASI module.
	The hardware can be used to perform random projection, PCA whitening, ICA, or a combination of random projection with the other two algorithms.
	This not only achieves implementing different dimensionality reduction algorithms on the same piece of hardware, but also allows dealing with higher number of dimensions.

	\section{Experimental Results And Discussion} \label{sec:experiments}
	
	\subsection{Dataset}
	In order to demonstrate the potentials of proposed solution, we use the Waveform Database Generator (Version 2) Dataset \cite{breiman1984classification}, which is publicly available on UCI Machine Learning Repository \cite{Lichman:2013}. 
	The dimensionality of input features is 40 where all features are noisy and the latter 19 are pure noise with a zero mean and variance of one. 
	The input features are all real numbers and there are no missing values in the dataset. 
	There are three classes of waves and the output classes represent combinations of two out of three of these base waves. 
	The number of samples is 5000 where we use the first 4000 for training of our models and the remaining 1000 for testing. 
	The objective is to classify samples into the three output classes. 
	In this work, we remove the latter eight input features and therefore, reduce the number of features that are pure noise to 13.
	As a result, the total number of input features will be 32.

	\subsection{Machine Learning Model}
	Our machine learning model consists of a dimensionality reduction module followed by an artificial neural network with two hidden layers and 64 neurons per each layer. 
	For dimensionality reduction, we use different algorithms such as EASI, random projection, or a combination of both. 
	For training the model, we first train the dimensionality reduction model in an unsupervised manner and reduce the dimensionality of input features. 
	After that, we train the neural network using features in the reduced space. 
	Finally, we use the dimensionality reduction model to decrease the dimensionality of test data and use the neural network for classification. 

	\subsection{Results}
	Table~\ref{table:software-results} compares the classification accuracy for different dimensionality reduction models and various number of intermediate and output features.
	It is observed that in configurations where the number of output features is the same, applying EASI independently or using random projection followed by EASI result in almost the same classification accuracy.
	However, as we will show later, the amount of hardware resources required for the latter is substantially smaller.

   \begin{table}[t]
		\centering
		\caption{Classification accuracy for different models and various number of intermediate and output features.}
		\begin{tabular}{c c c c c c c}
			$m$                & Algorithm 1        & $p$        & Algorithm 2           & $n$      & Accuracy (\%) \\
			\midrule
			32                 & --                 & --                                 & EASI              & 16                         & 84.6 \\
			32                 & Random Projection  & 24                                 & EASI              & 16                         & 84.5 \\
			\midrule
			32                 & --                 & --                                 & EASI              & 8                         & 80.9 \\
			32                 & Random Projection  & 16                                 & EASI              & 8                         & 80.8			
		\end{tabular}
		\label{table:software-results}
	\end{table}
	
	Table~\ref{table:hardware-results} summarizes the amount of resources required for implementing models where the number of input dimensions is 32 and the number of output dimensions is 8, after successful synthesis on FPGA.
	In both implementations, 32-bit floating-point variables and operations are used.
	The target FPGA is part of Arria 10 family which includes 427,200 adaptive logic modules (ALMs), 55,562,240 bits of block RAM, and 1518 DSP blocks. 
	It can be observed that the number of digital signal processors (DSPs), adaptive logic modules (ALMs), and bits required to store values in registers is reduced by a factor of two in the second scenario.
	In general, it is expected that the amount of savings will be proportional to $m / p$.
	As a result, using the random projection module to decrease the intermediate dimensionality further will lead to a more efficient hardware implementation.
	However, this typically affects the classification accuracy of different models.
	Therefore, the designer needs to trade off the hardware cost and accuracy in order to find a desirable point for number of intermediate features that achieves a relatively high classification accuracy, but reduces the hardware cost as much as possible.
	Note that the number of resources presented in Table~\ref{table:hardware-results} are more than the capacity of the target FPGA board and these numbers demonstrate the projected amount of required resources.

	Note that the pipelined implementation allows all algorithms to operate at the same clock frequency.
	As a result, using random projection followed by EASI does not lead to a lower frequency of operation, but slightly increases the latency.
	On our target FPGA, the post-place and route frequency of operation is 106.64\si{\mega\hertz}. 

   \begin{table}[t]
		\centering
		\caption{Comparison of hardware cost between EASI and random projection followed by EASI.}
		\begin{tabular}{c c c c c c}
			Input              & Intermediate       & Output            & DSPs      & ALMs        & Registers \\
			\midrule
			32                 & --                 & 8              & 4052     & 38122      & 138368 \\
			32                 & 16                 & 8	             & 2212     & 70031      & 75392 
		\end{tabular}
		\label{table:hardware-results}
	\end{table}

\balance

	\section{Conclusion} \label{sec:conclusion}
	In this work, we presented a hardware-friendly algorithm for improving the scalability of existing dimensionality reduction models.
	Additionally, we presented a reconfigurable hardware implementation that is capable of performing random projection, PCA whitening, and ICA through the EASI algorithm.
	The part of hardware implementation that improves scalability divides the whitening and rotation tasks between the random projection and EASI modules, respectively.
	This allows improving the hardware cost by a linear factor which is proportional to the ratio of the number of input features to intermediate features.
	Our experimental results show a 2x hardware cost reduction for a specific dataset, without affecting the classification accuracy by more than 0.1\%.

	\section*{Acknowledgements}
	This research was sponsored in part by contracts from DARPA's Microsystems Technology Office and the National Science Foundation.

	\bibliographystyle{IEEEtran}
	\bibliography{IEEEabrv,easi_scalable}
	
\end{document}